\documentclass[conference]{IEEEtran}
\IEEEoverridecommandlockouts
\usepackage{cite}
\usepackage{amsmath,amssymb,amsfonts}
\usepackage{algorithmic}
\usepackage{graphicx}
\usepackage{textcomp}
\usepackage{xcolor}
\usepackage{subfigure}
\def\BibTeX{{\rm B\kern-.05em{\sc i\kern-.025em b}\kern-.08em
    T\kern-.1667em\lower.7ex\hbox{E}\kern-.125emX}}
\begin{document}

\title{Feature matching in Ultrasound images\\
}

\author{\IEEEauthorblockN{Hang Zhu}
\IEEEauthorblockA{\textit{dept. Computer Science} \\
\textit{University of Alberta}\\
Multimedia\\
Edmonton, Canada \\
hzhu6@ualberta.ca}
\and
\IEEEauthorblockN{Zihao Wang}
\IEEEauthorblockA{\textit{dept. Computer Science} \\
\textit{University of Alberta}\\
Multimedia\\
Edmonton, Canada \\
zihao17@ualberta.ca}

}

\maketitle

\begin{abstract}

Feature matching is an important technique to identify a single object in different images. It helps machine to construct recognition of a specific object from multiple perspectives. For years, feature matching has been commonly used in various computer vision applications, like traffic surveillance, self-driving, and other systems. With the arise of Computer-Aided Diagnosis(CAD), the need of feature matching technique also emerges in medical imaging field. In this paper, we present a deep learning based method specially to ultrasound images. It will be examined against existing methods that have outstanding results on regular images. As the ultrasound images is different to regular images in many fields like texture, noise type, and dimension, traditional methods will be evaluated and optimized to be applied to ultrasound images. 

\end{abstract}


\section{Introduction}
\label{intro}
In clinic practice, ultrasound has become one of the most significant imaging modes. It is one of the most commonly used means of disease diagnosis because of its efficiency, low-cost, and convenience. However, due to the severe speckle noise, the poor image contrast and shadowing artifacts, noise reduction of ultrasonic images have always been an issue to be tackled in the digital image domain.\cite{Su2011} Additionally, the interpretation of ultrasound images highly depends on the doctor, which has a strong subjectivity. Consequently, computerized aided diagnosis systems are required in medical institutions to assist doctors in analyzing and interpreting ultrasound images to improve the accuracy, objectivity, and repeatability of examinations.
In recent years, the rise of deep learning algorithms has made great progress in the field of image processing and outperform their traditional counterparts ranging from feature extraction, image denoising, and other applications. 
However, the domain of ultrasound processing is still limited by several factors that are difficult to resolve. Firstly, the limited data set has become a bottleneck to the further application of deep learning methods in medical ultrasonic image analysis. It has been lectured that the flourish of neural network is in large part because of abundant data sets available\cite{ono2018lf}. Secondly, though some unsupervised learning methods can be trained end-to-end, some manual works are still indispensable in the training procedure, for instance, labeling. Whereas it could be extremely expensive to involve professionals, such as doctors to do such repetitive work. Last but not the least, a model trained from regular images will not be suitable for the single-channel, low contrast, and noise-filled ultrasound images. From those drawbacks mentioned above, there are stills technical problems for this project to resolve and explore.
The task consists of two parts: feature detection and feature matching(linking). Existing methods for feature detection, such as SIFT(Scale Invariant Feature Transform), SURF(Speeded Up Robust Feature), and BRISK (Binary Robust Invariant Scalable Keypoints) are mainly focused on regular images. Ultrasound images are slightly different to regular images since they usually suffer from the severe speckle noise, the poor image contrast and shadowing artifacts. Our goal is to examine and develop algorithms to discover an effective method for ultrasound images feature matching task.

\section{Related Work}
\label{RW}

\subsection{Review of Image Processing Based Methodology}
Since the 1970s, with the development of computer systems, the process of image processing became more effective and less money-consuming. Meanwhile, medical image analysis was performed by some traditional methodologies, for instance, edges and line detection filters, and also some mathematical morphological methods\cite{litjens2017survey}. The appearance of some powerful algorithms such as SIFT(Scale Invariant Feature Transform) also advanced the field of medical image\cite{Sun2015SIFT}.

\subsubsection{Organ segmentation}
Image processing methods took an important part in anatomy. Early researches on organ segmentation is a good example of it. External organs were firstly studied through regular images of the facial area. The research presented a feature detection method for nostrils and nose-side. With the result that over 90 percents of videos with no than 3 pixels shift of the target region, this energy-minimizing method shows great capability in nose detection\cite{yin2001nose}. Later in the field, more and more researches emerge with a powerful model and focus on more advanced use of image processing methods.

In 2007, Cheng et al.\cite{cheng2007airway} performed a study to extract and reconstruct a 3D airway model from Cone Beam CT data, then estimate the volume of the airway. The purpose of this project is to assist the tracking of the changes of the airway before and after a medical procedure. They proposed a method named gradient vector flow (GVF) snakes, which is a refined variant of snake/active contour model. The basic model finds the continuous edge of certain areas by converging a curve to the edges through iterations of energy minimization, and GVF snake model further improves the performance by eliminating the constraints of concave boundaries and initialization far from minimum. The essential core of the model can be summarized as minimizing the following energy function:

\begin{equation}
E = \iint \mu (u_x^{2}+u_y^{2}+v_x^{2}+v_y^{2}) + \left | \triangledown f \right |^{2}\left | v-\triangledown f \right |^{2}dxdy
\end{equation}
where $ f(x,y) $is the edge map of the image and $\mu$ is a tradeoff parameter.
There are some notable points in the project that reveal some limitations to the proposed model. In the project, the initialization of the active contour is still performed manually. Even though human interactivity is minimized to the first slice only, and the further transactions between following slices are inherited automatically, it still defines the procedure as semi-automatic rather than full-automatic. However, the method shows a very promising result considering the distance between CT slices. 

Energy-based segmentation methods show a reliable performance on the task, and it has evolved over the years. Recent research in 2019 on the segmentation topic proposed a level set method(LSM) driven by adaptive hybrid region-based energy\cite{han2019level}. The energy they defined is as followed:
\begin{equation}
E_{AH}=\rho (x)\cdot E_G(\varphi )+(1-\rho (x))\cdot E_L(\varphi)+E_R(\varphi)
\end{equation}
where $E_G$, $E_L$, and $E_R$ represent global region-based energy, local region-based energy, and regularized energy respectively. $\rho$ denotes the local intensity features. With this complex energy, the model can take advantage of both global and local-based regions and coping regardless of the homogeneity of the image. The proposed model is proven outperforming many existing LSM models in segmentation accuracy. As an important application in the medical imaging analysis field, image segmentation has revealed the capability of image processing in computer-aided diagnosis.

\subsubsection{Information-Theoretic approach}
Ultrasonography is also a commonly used medical image method for its cost-efficiency. However, computer-aided diagnosis of ultrasound images is usually challenging since they are
always corrupted by the speckle noise, which is distributed randomly, discretely and can also blur interesting features. The information-theoretic method mentioned in this paper\cite{4462338} provides a brand new idea for extracting features from speckled ultrasound images. Their work mainly focuses on extracting salient edges and interesting features embedded in speckles by comparing adjacent speckled areas horizontally and vertically. Unlike other feature detection methods, for instance, Canny edge detector\cite{4767851}, Derivative of Gaussian,  which are commonly used in edge detection, this algorithm is mainly focused on ultrasound images.

The detector in this article firstly use Rayleigh or Fisher-Tippett distributions to model the speckled regions, then deriving analytic expressions using the J-divergence to characterize differences.

\begin{figure*}[h]
  \centering
  \subfigure[Original image]{
  \begin{minipage}[t]{0.3\linewidth}
  \centering
  \includegraphics[width=2in]{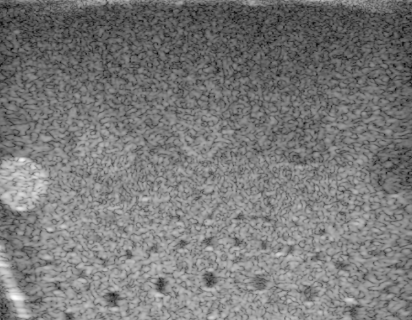}
  \end{minipage}
  }
  \subfigure[RayleighFisher-Tippett]{
  \begin{minipage}[t]{0.3\linewidth}
  \centering
  \includegraphics[width=2in]{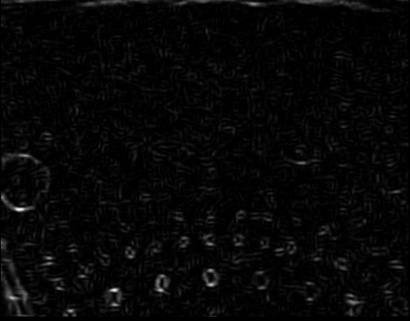}
  \end{minipage}
  }
  \subfigure[Canny edge detection]{
  \begin{minipage}[t]{0.3\linewidth}
  \centering
  \includegraphics[width=2in]{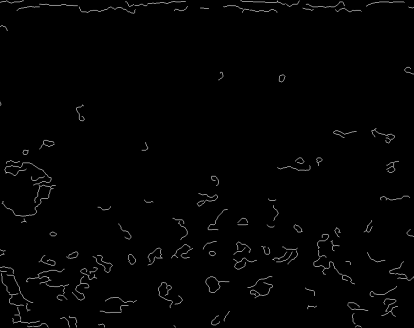}
  \end{minipage}
  }
  \caption{Information-Theoretic approach}\cite{4767851}
  \label{ITA}
\end{figure*}

Comparing with feature maps derived from image gradient operation, which still remained some speckled clusters, the information-theoretic method used a sliding window to eliminate the influence of noise. As the window size increased, more features were detected. More precisely, using J-divergence to measure differences between regions horizontally and vertically, functionally similar to gradient operation and generated a feature map(shows in \ref{ITA}), for more details of the derivation of formula, please refer to \cite{4767851}.
\begin{equation}
F_{j}(x, y)=\sqrt{J_{x}(x, y)^{2}+J_{y}(x, y)^{2}}
\end{equation}

\subsubsection{Feature detection and tracking}
Feature detection and tracking could be an easy task for a human, but this could be challenging for computer vision systems. This research field is hindered by several factors, for instance, shape characteristics of targets, noised-corrupted boundaries. This article\cite{log2005} mentioned an algorithm using Gaussian and Laplacian of Gaussian weighting functions to detect robust features. The main idea was to use weight functions, in more detail, to give more weights to regions more reliable to analyze. The Gaussian and LoG weight functions can be represented as follow.
\begin{equation}
_{\text {Gauss }}(x)=\left\{\begin{array}{ll}
\frac{1}{\sqrt{2 \pi \sigma^{2}}} \exp \left(-\frac{x^{2}}{2 \sigma^{2}}\right) & (x) \in \mathrm{FW} \\
0 & \text { elsewhere }
\end{array}\right.
\end{equation}

\begin{equation}
w_{\mathrm{LoG}}(x)=\left\{\begin{array}{ll}
\frac{1}{\sqrt{2 \pi \sigma^{2}}}\left[1-\frac{x^{2}}{\sigma^{2}}\right] \exp \left(-\frac{x^{2}}{2 \sigma^{2}}\right) & (x) \in \mathrm{FW} \\
0 & \text { elsewhere }
\end{array}\right.
\end{equation}

What is more, the Hough transform mentioned in this paper\cite{10.1016/S0167-8655(02)00159-9} could also be used for panoramic images feature tracking, which may be useful for processing panoramic medical images.

\subsubsection{Back Subtraction}
This paper\cite{inbook2} proposed a real-time background subtraction method called "Different clustering". Unlike traditional clustering algorithms that require several iterations to update kernels before convergence, this novel method only needs 2 clusters, meaning that the computational resources and the time required will decrease dramatically. Consequently, this method can be applied for video analysis. The main idea of this method is to extract difference vectors from target and background, and form a foreground mask after clustering and morphological processing. The improved clustering algorithm can be described as follow:
\begin{equation}
\mathcal{F}(x, y)=\underset{S}{\operatorname{argmin}} \sum_{i=1}^{k} \sum_{x, y \in S_{i}}\left\|D(x, y)-\mu_{i}\right\|^{2}
\end{equation}
Another method was raised by this paper\cite{rocket}, naming Robust Principal Components Analysis (RPCA) which was used to subtract the background from a freely moving camera. After optical flow was extracted, they implemented motion decomposition using RPCA. Finally integrated foreground masks.

Although the experiments took in these papers were based on RGB images, but the idea might be app-liable to separate the ultrasound feature map from noised background.

\subsubsection{GLCM and PCA}
This article\cite{inproceedings} proposed several approaches, including image feature extraction using Gray-level Co-occurrence Matrix(GLCM), feature selection using principal component analysis(PCA). The main idea of GLCM is describing the joint distribution of pixel intensity with a certain spatial position relation, by calculating and compute statistical distributions of gray values at different points to extract twenty-two texture-based features. After features being selected, we need to implement dimensionality reduction and feature selection to optimize computational complexity. The function we mentioned above may be used to optimize ultrasound feature matching in our project.

\subsubsection{Scale-invariant feature transform(SIFT)}
In the field of feature detection, SIFT is a state-of-art algorithm as its stable performance regardless of scaling, orientation, illumination change, affine distortion under certain extend\cite{lowe1999object}. With a comprehensive application example provided by the author to show the capability of the algorithm. The result turns out to be extraordinary. The example also demonstrates a standard procedure of feature detection and matching task: firstly using a feature descriptor to extract key points from pictures, then employing a matching algorithm to create links feature between images\cite{lowe2004distinctive}. The guidance is of great use to our project.

\subsection{Review of Learn-based Techniques}
Before 2010, the field of medical image analysis was dominated by systems that derive from machine learning algorithms and hand-crafted features. Examples include active shape models used for shape segmentation. These systems became significantly essential and paved the way for subsequent medical image processing applications for commercial use\cite{litjens2017survey}. Whereas the supervised techniques are hindered by the limited medical knowledge of computer scientists and limited programming skills of doctors.

\subsubsection{Feature extraction and classification with SVM}
Support Vector Machine(SVM) is a traditional supervised machine learning algorithm, which is widely used for classification. This paper\cite{yu2015lumbar} raises a method that contains feature extraction, feature matching and ultrasound image classifications. What worth mentioning is that during their study, they collected 40 ultrasound video streams and randomly captured images as training and testing data. In order to eliminate the speckle and noise generated by ultra wave, they implemented two normalized Gaussian filters. The authors also mentioned the limited feature dimension by using two to reduce local variance which can hinder optimally train. So, they use two methodologies to do dimensional reduction: template matching method to detect key points and midline detection to obtain features. The figure illustrates the procedure of this work.

Interestingly to point out, the co-worker of this article are two clinical experts, who are responsible for labelling the data, it is inevitable for them to have a disagreement in medical images. This may have an adverse impact on the accuracy of models. As mentioned in previous parts, hand-crafted features are not only induced by the high cost but also influenced by individual subjectivity. These factors may be the main reason for the underdevelopment in this domain, fortunately, unsupervised deep neural networks bring a new turning point.

\subsection{Review of Deep Neural Networks}
Logically, the next stage of development should be the computer extract features automatically and learn from themselves. The realization of this concept benefits from the advancement of computing power, which also results in the increasing utilization of Deep Learning. A successful algorithm in this domain is called Convolutional Neural Network(CNN). CNN contains many layers that apply a convolution filter to inputs to achieve dimensionality reduction.  During the gradual transition from systems that use handcrafted features to systems that learn features from data, the medical image analysis domain is rarely involved. 

\subsubsection{ Image segmentation with Deep Learning}
One of the basic applications of Deep Learning methods in Urology trains a generic U-Net model using publicly available CT scans. The motivation was to employ automatic techniques in the procedure of urologist’s decision making. The model shows great potential even without specific design accordingly\cite{graham2019accurate}. Though the performance is barely acceptable due to the lack of generalization, it still reveals a good capability of deep learning methods. The project also exposes overfitting could be an issue when models don't have prevention mechanisms.

Further research from Ji et al.in Ultrasound imaging field is to use a Fully Convolutional Network-based method to segment the different layers of blood vessels from IntraVascular UltraSound(IVUS) data\cite{yang2018ivus}. The same task was done manually by professionals, but with the rise of 3D surgical model reconstruction, the need for the data increases beyond human capability. The proposed model is called Dual-Path U-Net(DPU-Net) is a segmentation-purposed refinement of FNCs. The two major components of it firstly feed the input through an encoder network to generate a deep feature map of low resolution by image processing and downsampling. Then it restores the deep feature map back to the original through a decoder network. In the experiment, DPU-Net outperforms the general proposed method, namely SegNet and U-Net\cite{yang2018ivus}. As a deep architecture-based work, this well-designed model is proven to outperform existing conventional methods on distinguishing lumen and media vessel walls. The success of deep learning methods on image segmentation task provides a considerable option to our topic. In the progress of ultrasound image feature detection/matching, if the image could be preprocessed with such a method to extract a certain area of interest, there will be a boost in the accuracy, as one of the challenging problems is that it is hard to distinguish the noise and bone-texture.

\subsubsection{Image Recovery Network with Channel Attention Group(RCA-NET)}
The method raised in this article\cite{inbook1} provides a brand new sight of feature extraction while denoising. Their work was initially focused on recovery images from their foggy version using an end-to-end pipeline. Interestingly, they mentioned that traditional CNNs usually treated each channel equally, which led to the loss of varying information features and significant details. As a result, they proposed to use a channel-wise algorithm in each convolutional layer to retain more features.

They firstly obtained a trained model M(x) from the 'normal' end-to-end network, and then optimized M(x) gradually through a channel-wise recovery network in which structural details are drawn from to enhance images. Precisely, channel attention function was implemented in the following sequence: average pooling, convolution, ReLU function, convolution, Sigmoid function. The figure$a figure$ shows their work and we think this may be helpful when applying ultrasound image preprocessing.

\subsubsection{Deep-Learning Parkinson’s Disease Diagnosis}
Similarly, the method raised in this paper \cite{10.1007/978-3-030-54407-2_20} is used for ultrasound image processing. They built a model to diagnose PD by biomarkers instead of symptomatic diagnoses. More precisely, they constructed multiple models via a deep neural network to make a clear classification. After acquiring and preprocessing the data, they used Tenseorflow to build two distinct neural networks, one was three dimensions, while another is two dimensions. Interestingly, both models used "most-vote" to make the final classification. None of the models performed exceptional well(accuracy about 75\%).

\subsection{Review of Interpretability of Deep Neural Network}
This may be a new direction of research in the domain of artificial intelligence, but this methodology is quite inspiring not only to our work but also to other increasingly complex neural networks. The procedure of deep learning is known as a "black box", lack of implementation details of each layer makes it hard to interpret, which can hinder optimizations. More precisely, transparency and explanations are of great importance when scientists try  Implement optimizations as the intelligent system fails to produce satisfactory results. What is more, when a machine outperforms human experts, the goal of Interpretability is in machine teaching.\cite{selvaraju2017grad}

\subsubsection{Gradient-weighted Class Activation Mapping(Grad-CAM)}
\begin{figure*}[h]
  \centering
  \subfigure[Original image]{
  \begin{minipage}[t]{0.3\linewidth}
  \centering
  \includegraphics[width=2in]{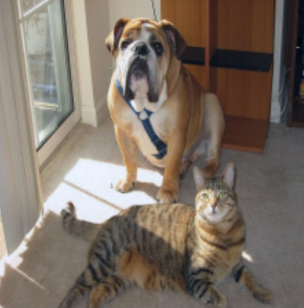}
  \end{minipage}
  }
  \subfigure[Grad-CAM ‘Cat’]{
  \begin{minipage}[t]{0.3\linewidth}
  \centering
  \includegraphics[width=2in]{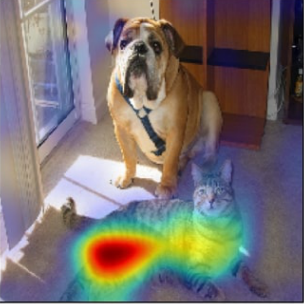}
  \end{minipage}
  }
  \subfigure[Guided Grad-CAM ‘Cat’]{
  \begin{minipage}[t]{0.3\linewidth}
  \centering
  \includegraphics[width=2in]{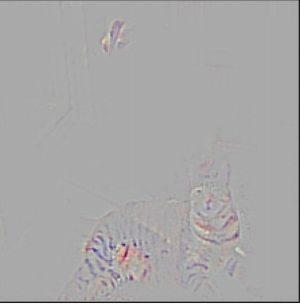}
  \end{minipage}
  }
  \caption{Grad-CAM approach}\cite{selvaraju2017grad}
  \label{CAM}
\end{figure*}

The method proposed in this paper\cite{selvaraju2017grad} uses the gradient of the target before the final convolutional layer to produce a localization feature to emphasize the target. Grad-CAM is applicable to various kinds of CNN without structural changes or re-training. This algorithm has a wide range of applications, but the reason it's mentioned here is that We think it can be used for feature extraction. Here is the output of Grad-CAM\ref{CAM}

\begin{figure*}[h]
    \centering
    \includegraphics[width=5in]{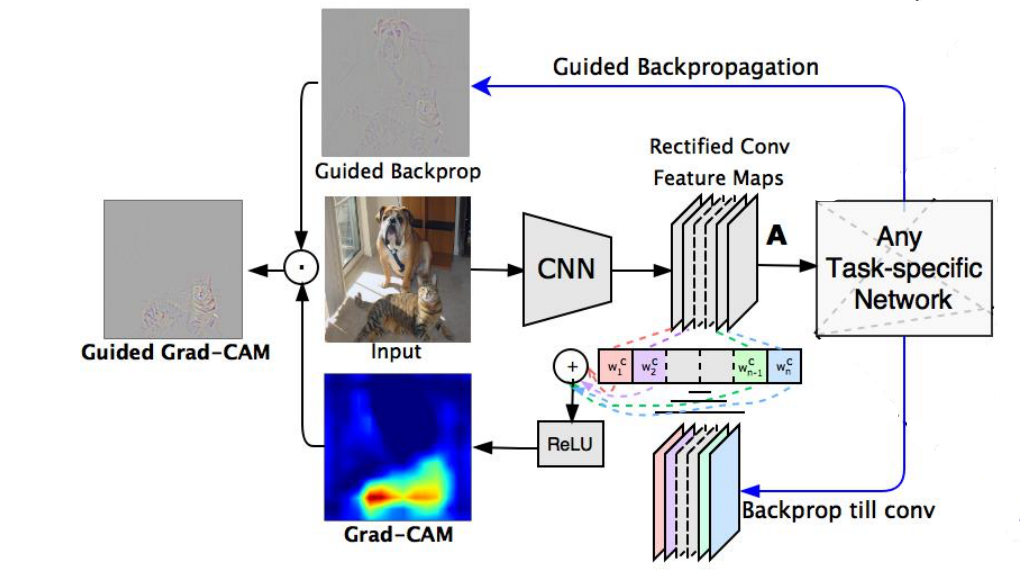}
    \caption{Procedures of Grad-CAM}
    \label{Grad-CAM}\cite{selvaraju2017grad}
\end{figure*}

Generally speaking, convolutional layers contains spatial information of images that will lose in the fully-connected layer. Logistically, the final convolutional layer will retain the best comprise between high-level semantics which cannot be comprehended by human eyes and spatial information. The algorithm can be mainly divided into two parts. Firstly heat map will be computed, which can reveal the relevant position of the target. Secondly, guided backpropagation which contains gradient will pointwise multiply with the output with the former step to generate guided Grad-CAM\ref{Grad-CAM}.

\subsection{Review of Data Set}
For deep learning methods, the performance is determined not only by the model architecture but also by the quality of the input data. Generally speaking, a larger data set will provides more choice of architectures. Due to the lack of data, we are considering adopting data augmentation in this project. The purpose of this method is to generate new data from the existing data set. For image data, shifting, rotating, and transforming have been commonly used. However, ultrasound images are special as it does not support certain strategy such as colour manipulations as it may change the nature of an ultrasound image.  Therefore, data augmenter need to be fully considered and examined. Meanwhile, data quality improvement will also be considered.
\subsubsection{Speckle noise}
As mentioned before, speckle noise is a major influencing factor in ultrasound data. Many approaches have been proposed to address this issue. A survey from Czerwinski et al. provides a comprehensive assessment of mainstream methods\cite{czerwinski1998line}. The survey concludes that the optimal operator is hard to achieve as the computational cost is too high. The linear detection outperforms the other sub-optimal operators. However, the performance is based on many assumptions, for instance, one assumption is that the noise is expected to be Gaussian and uncorrelated. In real-world examples, that assumption can hardly be satisfied. Thus the issue remains challenging.

\subsubsection{IVUS augmenter}
In the research of IVUS segmentation, ultrasound image quality is also hindered. To enrich the training set and best utilize the resources, a real-time augmenter was adopted. In their experiments, one important observation is that for IVUS images and many other medical images, since the target area is strictly located at the center of the area, thus transformations that lead to image shifting will be ineffective and even result in reducing in performance\cite{yang2019robust}. Apart from regular image augmentation methods, they also proposed an approach that takes noises into consideration. The strategy is to simulate the effect of shadow masks, side vessels, and bifurcation masks. With the different combinations of these three influence factors, a single image could have the capability to produce up to 8 variations. This approach could also inspire our method of data augmentation since our data set shares a similar feature/noise as theirs.

\subsubsection{Image fusion}
As mentioned previously, the ultrasound image datasets have limited size, in order to get better training models, image augmentation, for instance, image fusion which can combine images taken under different exposure or of different contrast is of great importance. Their first method was raised by this paper\cite{6392943} which is called "QoE-Based Multi-Exposure Fusion". The algorithm is based on the concept that each pixel of the fused image is influenced by two local-defined factors, they respectively are perceived local contrast and colour saturation, which are inserted into their hierarchical multivariate Gaussian conditional random field model which outperform other multi-exposure fusion.

Another method was raised in this article\cite{10.1007/978-3-030-54407-2_27}. The author implemented image fusion in the gradient domain. First of all, the decomposition model was used to decomposed target images into three layers. Then the pulse-coupled neural network (PCNN) was applied to reconstruct three fused layers. 

\clearpage

\clearpage


\bibliographystyle{IEEEtran}
\bibliography{egbib}
\end{document}